\pdfoutput=1

\documentclass[11pt]{article}

\usepackage[preprint]{acl}

\usepackage{times}
\usepackage{latexsym}

\usepackage[T1]{fontenc}

\usepackage[utf8]{inputenc}

\usepackage{microtype}
\usepackage{amsmath}
\usepackage{amssymb}

\usepackage{inconsolata}

\usepackage{graphicx}
\usepackage{subcaption}  
\usepackage{float} 
\usepackage{tikz}
\usepackage[edges]{forest}
\usepackage{xcolor}

\title{A Survey of Retentive Network}

\author{
        Haiqi Yang$^{1}$ \quad Zhiyuan Li$^{1}$ \quad Yi Chang$^{1,2,3}$ \quad Yuan Wu$^{1}$\footnotemark[1] \\
        $^{1}$School of Artificial Intelligence, Jilin University \\
        $^{2}$Engineering Research Center of Knowledge-Driven Human-Machine Intelligence, MOE, China \\
        $^{3}$International Center of Future Science, Jilin University\\
        \{yanghaiqi24, zhiyuanl24\}@mails.jlu.edu.cn, \\
        yichang@jlu.edu.cn, yuanwu@jlu.edu.cn \\   
}

\begin{document}
\maketitle
\renewcommand{\thefootnote}{\fnsymbol{footnote}}
\footnotetext[1]{Corresponding authors}
\begin{abstract}
Retentive Network (RetNet) represents a significant advancement in neural network architecture, offering an efficient alternative to the Transformer. While Transformers rely on self-attention to model dependencies, they suffer from high memory costs and limited scalability when handling long sequences due to their quadratic complexity. To mitigate these limitations, RetNet introduces a retention mechanism that unifies the inductive bias of recurrence with the global dependency modeling of attention. This mechanism enables linear-time inference, facilitates efficient modeling of extended contexts, and remains compatible with fully parallelizable training pipelines. RetNet has garnered significant research interest due to its consistently demonstrated cross-domain effectiveness, achieving robust performance across machine learning paradigms including natural language processing, speech recognition, and time-series analysis. However, a comprehensive review of RetNet is still missing from the current literature. This paper aims to fill that gap by offering the first detailed survey of the RetNet architecture, its key innovations, and its diverse applications. We also explore the main challenges associated with RetNet and propose future research directions to support its continued advancement in both academic research and practical deployment.
\end{abstract}

\section{Introduction}
\label{sec:intro}
\citet{vaswani2017attention} proposed the Transformer architecture, which relies solely on the self-attention mechanisms. Owing to its ability to model long-range dependencies and its high degree of parallelism, the Transformer has emerged as the dominant paradigm in natural language processing (NLP). Beyond NLP, the Transformer has been successfully applied to a wide range of domains such as computer vision (CV), speech, and scientific areas like chemistry and bioinformatics, reflecting its versatility in modeling complex, long-range dependencies across modalities.
Despite its strengths, the Transformer architecture faces notable limitations. During training, its quadratic time complexity makes modeling long sequences computationally costly. In the inference phase, linear memory complexity arises from storing KV cache for each token, resulting in significant memory overhead.
Although various approaches have been explored to mitigate the complexity of the Transformer, achieving substantial reductions in computational overhead remains challenging\cite{choromanski2020rethinking,katharopoulos2020transformers,wang2020linformer}.

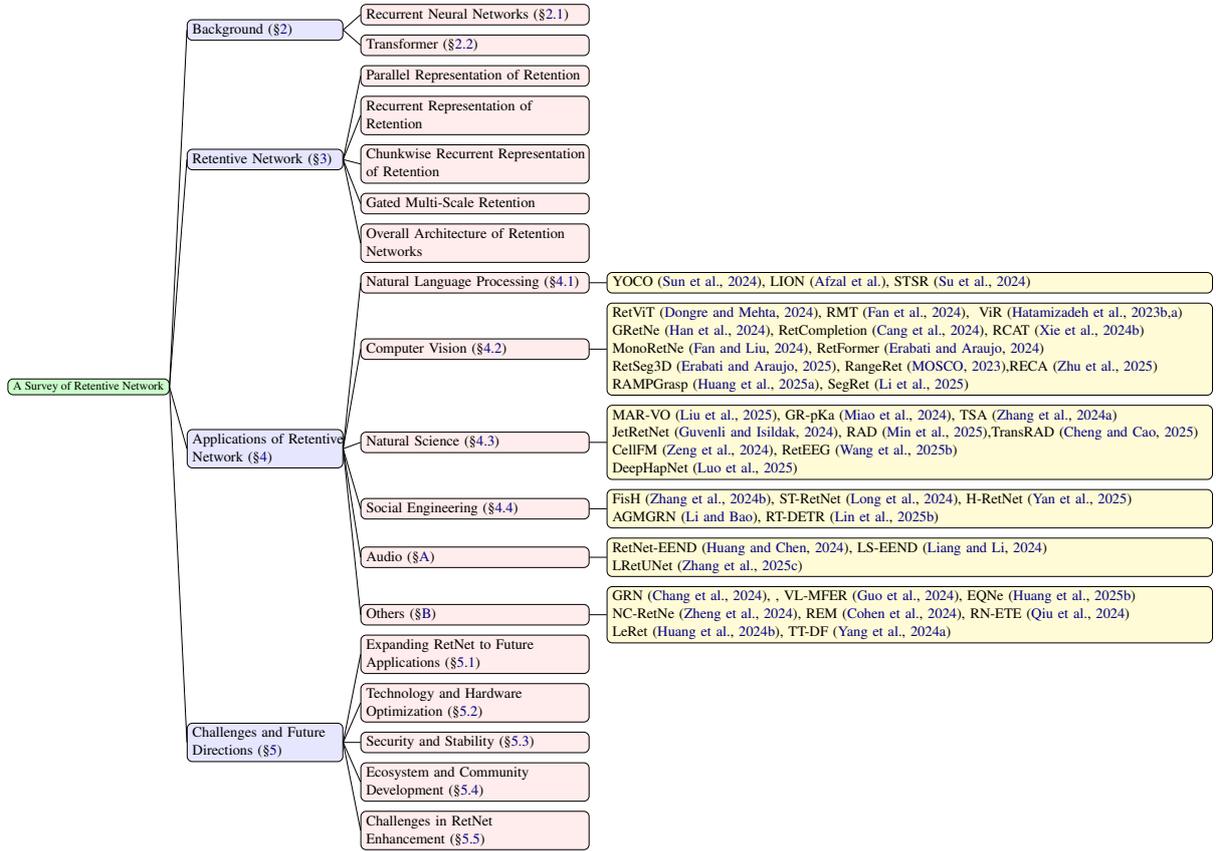
\begin{figure*}[t!]
    \vspace{-1.0cm}
    \centering
    \resizebox{\textwidth}{!}{

\begin{forest}
  for tree={
  grow=east,
  reversed=true,
  anchor=base west,
  parent anchor=east,
  child anchor=west,
  base=left,
  font=\small,
  rectangle,
  draw,
  rounded corners,align=left,
  minimum width=5em,
  inner xsep=4pt,
  inner ysep=1pt,
  },
  where level=1{text width=10em, font=\normalsize,fill=blue!10}{},
  where level=2{text width=15em, font=\normalsize,fill=pink!30}{},
  where level=3{text width=41em, font=\normalsize,fill=yellow!20}{},
  [A Survey of Retentive Network, fill=green!20
        [Background (\S \ref{sec:background})
            [Recurrent Neural Networks (\S \ref{sec:rnn})]
            [Transformer (\S \ref{sec:transformer})]
        ]    
        [Retentive Network (\S \ref{sec:retnet})
            [Parallel Representation of Retention]
            [Recurrent Representation of \\ Retention]
            [Chunkwise Recurrent Representation \\ of Retention]
            [Gated Multi-Scale Retention]
            [Overall Architecture of Retention \\ Networks]
        ]
        [Applications of Retentive \\ Network (\S \ref{sec:application})
            [Natural Language Processing  (\S \ref{sec:nlp})
            [YOCO \cite{sun2024you}{, }LION \cite{afzallion}{, }STSR~\cite{su2024sequence}]
            ]
            [Computer Vision (\S \ref{sec:cv}) 
            [RetViT~\cite{dongre2024retvit}{, }RMT~\cite{fan2024rmt}{, }
            ViR~\cite{hatamizadeh2023vira,hatamizadeh2023virb}
            \\ GRetNe~\cite{han2024gretnet}{, }RetCompletion~\cite{cang2024retcompletion}{, }RCAT~\cite{xie2024rcat}
            \\ MonoRetNe~\cite{fan2024monoretnet}{, }RetFormer~\cite{erabati2024retformer}
            \\ RetSeg3D~\cite{erabati2025retseg3d}{, }RangeRet~\cite{mosco2023exploiting}{,}RECA~\cite{zhu2025retention} \\ RAMPGrasp~\cite{huang2025rampgrasp}{, }SegRet~\cite{li2025segret}]
            ]
            [Natural Science (\S \ref{sec:n_s}) 
            [MAR-VO~\cite{liu2025mar}{, }GR-pKa~\cite{miao2024gr}{, }TSA~\cite{zhang2024transient} \\ JetRetNet~\cite{guvenli2024b}{, }RAD~\cite{min2025retentive}{,}TransRAD~\cite{cheng2025transrad} \\ CellFM~\cite{zeng2024cellfm}{, }RetEEG~\cite{wang2025convolutional} \\ DeepHapNet~\cite{luo2025deephapnet}]
            ]   
            [Social Engineering (\S \ref{sec:s_e}) 
            [FisH~\cite{zhang2024fast}{, }ST-RetNet~\cite{long2024st}{, }H-RetNet~\cite{yan2025multimodal} \\ AGMGRN~\cite{li5170149adaptive}{, }RT-DETR~\cite{lin2025disease}]
            ]
            [Audio (\S \ref{audio}) 
            [RetNet-EEND~\cite{huang2024long}{, }LS-EEND~\cite{liang2024ls}
             \\ LRetUNet~\cite{zhang2025lretunet}]         
            ]
            [Others (\S \ref{others}) 
            [GRN~\cite{chang2024graph}{, }{, }VL-MFER~\cite{guo2024vl}{, }EQNe~\cite{huang2025boost}
            \\ NC-RetNe~\cite{zheng20243d}{, }REM~\cite{cohen2024improving}{, }RN-ETE~\cite{qiu2024rn}
            \\ LeRet~\cite{huang2024leret}{, }TT-DF~\cite{yang2024tt}]
            ]
        ]
        [Challenges and Future \\ Directions (\S \ref{sec:challege_future})
            [Expanding RetNet to Future \\ Applications (\S \ref{sec:expanding})]
            [Technology and Hardware \\ Optimization (\S \ref{sec:technology})]
            [Security and Stability (\S \ref{sec:security})]
            [Ecosystem and Community \\ Development (\S \ref{sec:ecd})]
            [Challenges in RetNet \\ Enhancement (\S \ref{sec:cre})]
        ]
    ]
\end{forest}

    }
    \caption{Structure of this paper.}
    \label{structure}
\end{figure*}

To address the computational limitations of traditional Transformers, many research advances have emerged. Gated linear recurrent neural networks  \cite{qin2023hierarchically,de2402griffin} incorporated gating mechanisms to reduce the quadratic time complexity typically associated with Transformer training. State Space Models compressed sequence data into fixed-size representations, effectively mitigating the scaling issues inherent in Transformers \cite{gu2021efficiently,gu2023mamba}. Linear Transformers \cite{katharopoulos2020transformers} further alleviated memory and computational overhead by employing linear attention mechanisms, allowing both time and memory complexity to scale linearly with sequence length. The Receptance Weighted Key Value (RWKV) leverages linear attention to reduce computational complexity and memory usage during inference~\citep{peng2023rwkv,li2024survey}.
Among these, RetNet \cite{sun2023retentive} stands out as a compelling solution, it integrated a multi-scale retention mechanism which employs three computational paradigms namely parallel, recurrent, and chunkwise recurrent representations. By leveraging these paradigms, RetNet achieves performance comparable to Transformers while enabling constant-time O(1) inference, reduced memory overhead, and efficient long-sequence modeling.

By employing the retention mechanism, the decay mask makes RetNet very versatile for a wide range of applications, from NLP \cite{cheng2024pre}, CV \cite{fan2024rmt}, natural science \cite{luo2025deephapnet} to social engineering \cite{yan2025multimodal}.
With the rapid expansion of research and applications of RetNet, this survey aims to shed light on current progress in this field.
As depicted in Figure \ref{structure}, the remainder of this paper is organized as follows: Section \ref{sec:background} provides a systematic review of basic concepts, including RNN and Transformer architectures, Section \ref{sec:retnet} delves into the principle and mechanism of RetNet, and Section \ref{sec:application} explores the extensive applications of RetNet in diverse domains, including NLP, 
CV, natural sciences, social engineering, and audio processing. Section \ref{sec:challege_future} examines the primary challenges confronting RetNet and outlines prospective directions for future research.


\section{Background}
\label{sec:background}
\subsection{Recurrent Neural Networks}
\label{sec:rnn}
Recurrent neural networks (RNNs) are capable of learning features and long-term dependencies from sequential and time-series data \cite{salehinejad2017recent}. Specifically, RNN introduced a recurrent architecture that maintains a hidden state, enabling the modeling of sequential data with variable lengths through shared weights across time steps \cite{hochreiter1997long}. 
The operational mechanism of RNN is captured by the following mathematical formulation:
\begin{equation}
h_{t} = f_{H}(W_{hh} \cdot h_{t-1} + W_{hx} \cdot x_t + b_h)
\label{eq:1}
\end{equation}
\begin{equation}
y_{t} = f_{O}(W_{ho}\cdot{h}_{t}+b_{o})
\label{eq:2}
\end{equation}
where $h_t$ denotes the hidden state at time step $t$, and $x_t$ is the input at time $t$. The function $f_H(\cdot)$ is the hidden layer activation function, and $f_O(\cdot)$ is the output activation function. $y_t$ denotes the output at time $t$. $W_{hh}$, $W_{hx}$, and $W_{ho}$ are the weight matrices connecting the hidden-to-hidden, input-to-hidden, and hidden-to-output layers, respectively. $b_h$ and $b_o$ are the bias vectors for the hidden and output layers.

Despite the RNN's strength in modeling temporal sequences, a major limitation is the vanishing gradient problem, which causes gradients to decay exponentially over time steps. This significantly impairs the network's ability to retain and utilize information from distant past inputs~\cite{bengio1994learning}.

To mitigate the challenges of vanishing or exploding gradients encountered by RNN when processing extended sequences, researchers have developed several advanced variants. Long short-term memory network (LSTM) \cite{hochreiter1997long} incorporates gating mechanisms to regulate information flow, enabling effective capture of long-term dependencies.
Gated recurrent unit (GRU) \cite{cho2014properties} offers a simplified architecture compared to LSTM while delivering comparable performance. Bidirectional recurrent neural network (Bi-RNN) \cite{schuster1997bidirectional} processes sequences in both forward and reverse directions simultaneously, providing a more comprehensive understanding of sequential patterns.
\subsection{Transformer}
\label{sec:transformer}
The Transformer architecture dispenses with recurrence entirely, relying instead on self-attention mechanisms to model global dependencies between inputs and outputs~\cite{vaswani2017attention}. This fundamental shift enables the model to more effectively capture long-range relationships and supports stable, efficient training without the gradient propagation issues commonly associated with recurrent structures.

The attention mechanism enables models to capture dependencies among different positions within an input sequence, which is essential for learning contextual relationships. In Self-Attention, the input sequence is represented as a matrix \( X \in \mathbb{R}^{n \times d} \), where \( n \) is the number of tokens and \( d \) is the dimensionality of each token embedding. To generate the necessary attention components, the Transformer applies three independent trainable linear transformations to the input: the matrix \( X \) is projected into the query, key, and value spaces using the weight matrices \( W^Q \in \mathbb{R}^{d \times d_q} \), \( W^K \in \mathbb{R}^{d \times d_k} \), and \( W^V \in \mathbb{R}^{d \times d_v} \). As a result, we obtain \( Q = X W^Q \), \( K = X W^K \), and \( V = X W^V \).

Each row in \( Q \), \( K \), and \( V \) corresponds to a token in the sequence, where \( Q \) represents the queries used to attend to other tokens, \( K \) represents the keys that determine relevance, and \( V \) carries the actual content of the tokens. The Self-Attention output is computed using the scaled dot-product attention mechanism:
\begin{align}
\text{Attention}(Q, K, V) = \text{softmax}\left(\frac{QK^\top}{\sqrt{d_k}}\right)V ,
\end{align}
where \( d_k \) denotes the dimensionality of the key vectors. The scaling factor \( \sqrt{d_k} \) is used to mitigate the impact of large dot-product values, ensuring stable gradients and more effective learning.

The self-attention mechanism in the Transformer is extended to multiple attention heads, each capable of learning distinct attention weights to effectively capture diverse relational patterns. Multi-head attention enables the model to process different informational subspaces in parallel, enhancing its representational capacity.
\begin{equation}
head_i = Attention(QW_i^Q, KW_i^K, VW_i^V)
\end{equation}
\begin{equation}
\begin{aligned}
& MultiHead(Q, K, V) \\
&=Concat(head_1, \dots,head_h) W^O
\end{aligned}
\end{equation}
where $h$ denotes the number of attention heads, $Concat$ represents the concatenation operation, and $W^O$ is the trainable projection matrix, $W_i^Q$, $W_i^K$, and $W_i^V$ are the parameter matrice. Multi-head attention significantly enhances the Transformer’s capability to address NLP tasks and other forms of sequential data processing with improved efficiency and expressiveness.

\section{Retentive Network}
\label{sec:retnet}
\begin{figure*}[t]
  \centering
  \begin{subfigure}[b]{0.48\textwidth}
    \centering
    \includegraphics[height=6cm]{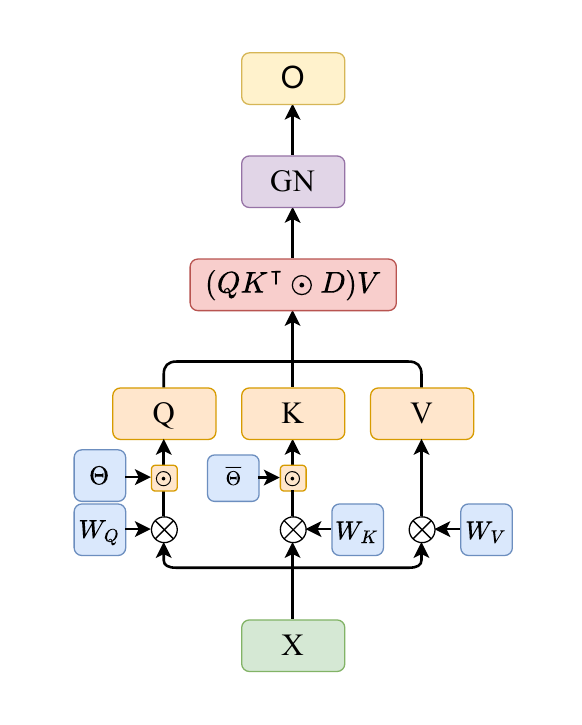}
    \caption{Parallel representation.}
    \label{fig:parallel}
  \end{subfigure}
  \hfill
  \begin{subfigure}[b]{0.48\textwidth}
    \centering
    \includegraphics[height=6cm]{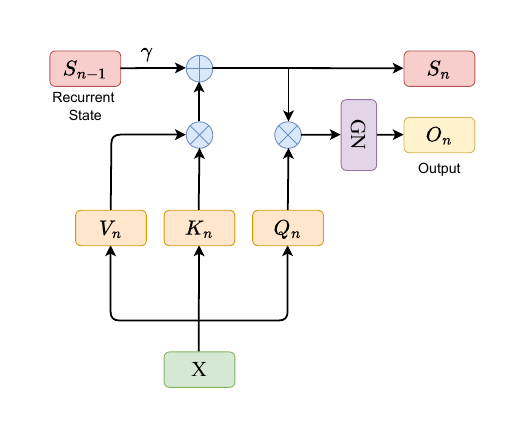}
    \caption{Recurrent representation.}
    \label{fig:recurrent}
  \end{subfigure}
  \caption{Dual form of RetNet. ``GN'' denotes GroupNorm.}
  \label{fig:arch}
\end{figure*}
RNNs have difficulty capturing long-range dependencies due to the vanishing gradient problem and their inherently sequential structure, which also limits parallelism~\cite{yu2019review}. Transformer, while effective at capturing long-range dependencies, face high computational complexity and inefficiency in processing long sequences~\cite{lin2022survey}. 
RetNet\cite{sun2023retentive} theoretically derived the connection between recurrence and attention and  proposed retention mechanism for sequence modeling. RetNet has been shown to achieve low-cost inference, efficient long-sequence modelling, Transformer-comparable performance, and parallel model training simultaneously.

RetNet is constructed as a stack of $L$ identical blocks, each comprising two core components: a Multi-Scale Retention (MSR) module and a Feed-Forward Network (FFN) module. For a given sequence of input \( x = x_{1} \cdots x_{|j|} \), where \( |j| \) represents the length of the sequence, RetNet utilizes an autoregressive encoding method to process the sequence. The input is packed into \( X^0 = [\textbf{x}_{1}, \cdots, \textbf{x}_{|j|}] \in \mathbb{R}^{|j| \times d_{\text{model}}} \), where \( d_{\text{model}} \) is the dimension of the hidden layer. Then compute the contextualized vector representations as follows:
\begin{equation}
  X^l = \mathrm{RetNet}_{l}(X^{l-1}), l\in [1, L].
\end{equation}

Retention mechanism with a dual form of recursion and parallelism is the key to the success of RetNet. Project the input $X \in \mathbb{R}^{|j|\times d_\text{model}}$ to $v_n = X_n \cdot w_{v}$, where $w_{v}$ is the trainable matrix that maps inputs to value vectors.
Then make the projection $Q, K$:
\begin{equation}
\label{eq:content:aware}
Q = X W^Q, \quad K = X W^K,
\end{equation}
where $W^Q, W^K \in \mathbb{R}^{d\times d}$ are learnable matrices.

\begin{figure}[t]
    \centering
    \includegraphics[width=0.5\textwidth]{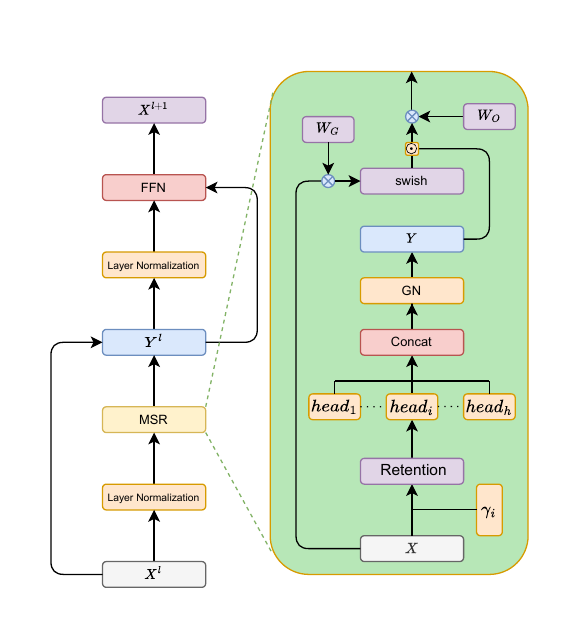}
    \caption{Overall architecture of RetNet.}
    \label{fig:overall}
\end{figure}

Consider a sequence modeling problem, through the state $\textbf{s}_{n}\in\mathbb{R}^{d\times d}$ mapping $v_n$ to a vector of $o_n$. 
\begin{equation}
\begin{aligned}
\label{eq:7}
&\mathbf{s}_{n} = A\mathbf{s}_{n-1} + K_n^\top v_n \\
&o_n = Q_n\mathbf{s}_{n} = \sum_{m=1}^n Q_n A^{n - m} K_m^\top v_m
\end{aligned}
\end{equation}
where $K_n, Q_n$ is the projection of the time step n.

Further, diagonalize $A = \Lambda(\gamma e^{i \theta})\Lambda^{-1}$, where $\Lambda$ is the reversible matrix,  $\gamma$ is the decay mask, according to Euler's formula $e^{i\theta}=[\cos\theta_1,\sin\theta_2,\cdots,\cos\theta_{d-1},\sin\theta_d]$, then $A^{n - m} = \Lambda(\gamma e^{i \theta})^{n-m}\Lambda^{-1}$, $n$, $m$ is the time step. Equation \ref{eq:7} becomes:
\begin{equation}
\begin{aligned}
o_n &=\sum_{m=1}^n(Q_n(\gamma e^{i\theta})^n)(K_m(\gamma e^{i\theta})^{-m})^\top v_m \\
&= \sum_{m=1}^n \gamma^{n-m} (Q_n e^{in\theta})(K_m e^{im\theta})^\dag v_m
\end{aligned}
\end{equation}
where $Q_n(\gamma e^{i\theta})^n$, $K_m(\gamma e^{i\theta})^{-m}$ is the xPos~\cite{sun2022length}, $^\dag$ is the conjugate transpose. $e^{in\theta}$ and $e^{im\theta}$ serve as rotational factors that encode positional information using complex exponential forms, where $\theta$ denotes the learnable parameters employed to model relative phase differences for the purpose of capturing sequential dependencies.

\paragraph{Parallel Representation of Retention}
As shown in Figure \ref{fig:parallel}, the retention layer is defined as:
\begin{equation}
\begin{aligned}
& Q = (X W^Q) \odot \Theta, \quad K = (X W^K) \odot \overline{\Theta}, \\
& V = X W^V, \\
& D_{nm} = 
\left\{
\begin{aligned}
& \gamma^{n-m}, &n \ge m \\
& 0, &n < m \\
\end{aligned}
\right., \\
& \mathrm{Retention}(X) = (Q K^\top \odot D)V
\end{aligned}
\label{eq:ret:parallel}
\end{equation}
where $\odot$ is the Hadamard product, $\Theta$ is the position-dependent modulation term, and $\overline{\Theta}$ denotes its complex conjugate, and $D \in \mathbb{R}^{|j|\times |j|}$ constitutes a unified matrix that jointly encodes causal masking and exponential decay as a function of relative positional distance.

\paragraph{Recurrent Representation of Retention}
As shown in Figure \ref{fig:recurrent}, at the $n$-th timestep, the output is recurrently obtained as follows:
\begin{equation}
\begin{aligned}
\label{eq:ret:recurrent}
&S_n = \gamma S_{n-1} + K_n^{\top} V_n \\
&\mathrm{Rete}\mathrm{ntion} (X_n) = Q_n S_n, n = 1, \cdots, |j| \\
\end{aligned}
\end{equation}

\paragraph{Chunkwise Recurrent Representation of Retention}
The input sequences are segmented into chunks. Within each chunk, the computation is carried out using the parallel representation Equation \ref{eq:ret:parallel}. In contrast, information across chunks is propagated using the recurrent representation Equation \ref{eq:ret:recurrent}. Specifically, let $B$ denote the chunk length. The retention output of the $i$-th chunk is computed as follows:
\begin{equation}
\begin{aligned}
&Q_{[i]} = Q_{Bi:B(i+1)}, \\
&K_{[i]} = K_{Bi:B(i+1)},\\
&V_{[i]} = V_{Bi:B(i+1)}, \\
&R_{i} = K_{[i]}^\top (V_{[i]} \odot \zeta) + \gamma^{B} R_{i-1},\\
&\mathrm{Retention}(X_{[i]}) = \underbrace{(Q_{[i]} K^\intercal_{[i]} \odot D) V_{[i]}}_{\text{Inner-Chunk}}\\
&\hspace{6.7em} + \underbrace{(Q_{[i]} R_{i-1}) \odot \xi}_{\text{Cross-Chunk}}\\
&\xi_{ij} = \gamma^{i+1},\quad \zeta_{ij} = \gamma^{B-i-1}
\end{aligned}
\end{equation}
where ${[i]}$ indicates the $i$-th chunk, i.e., $x_{[i]} = [x_{(i-1)B+1} , \cdots , x_{iB}]$. $\zeta$ and $\xi$ are exponential decay factors that modulate the influence of intra-chunk and inter-chunk information.

\paragraph{Gated Multi-Scale Retention}
In each layer, the number of retention heads is defined as $h = d_{\text{model}} / d$, where $d$ denotes the head dimension. Each head is associated with distinct parameter matrices $W^Q, W^K, W^V \in \mathbb{R}^{d \times d}$. MSR mechanism assigns a unique decay factor $\gamma$ to each head. For simplicity, identical $\gamma$ values are used across different layers and kept fixed.
To enhance the non-linearity of the retention layers, a swish gate~\cite{hendrycks2016gaussian,ramachandran2017swish} is introduced. Given the input $X$, the computation of the layer is defined as follows:
\begin{equation}
\begin{aligned}
\label{eq:msr}
&\mathbf{\gamma} = 1 - 2^{-5-\mathrm{arange}(0, h)} \in \mathbb{R}^{h} \\
&\mathrm{head}_i = \mathrm{Retention}(X, \gamma_i) \\
&Y = \mathrm{GN}_{h}( \mathrm{Concat}(\mathrm{head}_1, \cdots, \mathrm{head}_h) ) \\
&\mathrm{MSR}(X) = (\mathrm{swish}(X W^G) \odot Y) W^O
\end{aligned}
\end{equation}
where $W^G$, $W^O\in \mathbb{R}^{d_{\text{model}} \times d_{\text{model}}}$ are learnable parameter matrices. $\text{arange}(0, h)$ denotes a vector of integers from $0$ to $h-1$, used to assign distinct decay scales across $h$ attention heads. $\mathrm{GN}$ denotes Group Normalization~\cite{wu2018group}, applied to each head output following the SubLN strategy in~\cite{shoeybi2019megatron}. Since each head employs a distinct $\gamma$ scale, their output variances differ, which necessitates separate normalization.

\paragraph{Overall Architecture of Retention Networks}
As illustrated in Figure~\ref{fig:overall}, an $L$-layer retention network is constructed by stacking MSR and FFN modules. The input sequence $\{x_i\}_{i=1}^{|j|}$ is first mapped to vector representations via a word embedding layer. The resulting embeddings, denoted as $X_0 = [x_1, \cdots, x_{|j|}] \in \mathbb{R}^{|j| \times d_{\text{model}}}$, serve as the initial input to the model. The final output is represented as $X^L$.
\begin{equation}
\begin{aligned}
\label{eq:arch}
Y^l &= \mathrm{MSR}(\mathrm{LN}(X^l)) + X^l \\
X^{l+1} &= \mathrm{FFN}(\mathrm{LN}(Y^l)) + Y^l
\end{aligned}
\end{equation}
where $\mathrm{LN}(\cdot)$ denotes the Layer Normalization function~\cite{ba2016layer}. The feed-forward network (FFN) is defined as
\[
\mathrm{FFN}(X) = \mathrm{gelu}(X W_1) W_2,
\]
where \(W_1\) and \(W_2\) are learnable parameter matrices, and \(\mathrm{gelu}(\cdot)\) is the Gaussian Error Linear Unit activation function.

\section{Applications of Retentive Network}
\label{sec:application}
\subsection{Natural Language Processing}
\label{sec:nlp}
RetNet has proven to be highly effective in a variety of NLP tasks due to its efficient retention mechanism. In language modeling, the decoder-decoder architecture with gated retention mechanism was introduced by \citet{sun2024you} to improve contextual understanding. 
For knowledge graph reasoning, \citet{cheng2024pre} utilized RetNet as an encoder. In multi-hop reasoning tasks, the STSR model presented by \citet{su2024sequence} employed RetNet's parallel retention module to speed up training and enhance performance in sequence-to-sequence reasoning tasks.
The LION framework, developed by \citet{afzallion}, adapted RetNet for bidirectional language tasks, incorporating fixed decay masks to efficiently capture long-range dependencies and reduce computational costs. \citet{afzal2025linear} further refined this idea in LION-D, a bidirectional variant of RetNet that supports linear-time inference while preserving the efficiency of parallel training. \citet{he2024densemamba} introduced DenseRetNet, which improves feature extraction by integrating dense hidden connections.

\subsection{Computer Vision}
\label{sec:cv}
RetNet and their variants have demonstrated broad applicability across various CV domains. 
The fixed decay mask enables RetNet to efficiently capture long-range spatial or temporal dependencies in images or videos.

\paragraph{Image Tasks.} 
RetViT replaces standard attention with parallelizable retention blocks to accelerate training while maintaining representational capacity \cite{dongre2024retvit}. \citet{fan2024rmt} extend RetNet's one-dimensional unidirectional decay matrix to a two-dimensional bidirectional decay matrix, thereby designing Manhattan Self-Attention (MaSA). ViR explores efficient vision backbones by redesigning the retention mechanism to support both parallel training and recurrent inference \cite{hatamizadeh2023vira}. Another ViR model leverages RetNet's block structure and multi-scale design to recursively capture contextual dependencies across spatial scales \cite{hatamizadeh2023virb}.
\citet{hu2024shifted} proposed the SwiFTeR architecture, which employs the Retention mechanism in the fusion model's decoder. SegRet applies multi-scale retention modules to strengthen hierarchical feature aggregation, boosting semantic segmentation accuracy \cite{li2025segret}.
Retention mechanism helps hyperspectral models reduce memory cost while preserving spectral discriminability \cite{arya2025efficient,paheding2024hyperspectral}.
GRetNet enhances spatial feature modeling via Gaussian-decayed retention based on Manhattan distance \cite{han2024gretnet}.
Incorporating MaSA into LoFTR-like frameworks improves coarse feature matching in challenging correspondence tasks \cite{sui2024efficient}.
Multi-focus image fusion benefits from bidirectional 2D retention that captures local spatial consistency \cite{huang2024efficient}.
RetCompletion applies a fast parallelized retentive decoder for real-time image inpainting \cite{cang2024retcompletion}.
The Cross-Axis Transformer integrates RetNet's recurrent retention mechanism to process visual attention across chunked image regions~\cite{erickson2023crossaxis}.
The RetNet-based retention module is cleverly applied to rotating target detection \cite{liu2024lightweight}.

\paragraph{Video Tasks.} RCAT combines RetNet with CLIP adapters, yielding strong results in video recognition across different datasets \cite{xie2024rcat}.
Maskable RetNet introduces learnable masking strategies, improving temporal localization in moment retrieval \cite{hu2024maskable}.
MonoRetNet proposed a half-duplex bidirectional retention design for monocular depth prediction from sequential frames \cite{fan2024monoretnet}.

\paragraph{3D Data Modeling.}  RetFormer incorporates spatial retention module tailored to 3D Transformer backbones \cite{erabati2024retformer}.
LION models point cloud sequences with linear-time complexity by applying groupwise retention in RNN-style architectures \cite{liu2024lion}.
RetSeg3D extends the retention concept from one-dimensional sequences to 3D voxel grids for improved semantic parsing \cite{erabati2025retseg3d}.
RangeRet introduces a Manhattan distance-based spatial decay, enhancing context aggregation in LiDAR segmentation tasks \cite{mosco2023exploiting}.
Octree-Retention Fusion exploits parallel retention with exponential decay masks to improve hierarchical context modeling in point cloud compression \cite{zhang2024octree}.

\paragraph{Cross-modal Tasks.} RECA refines multi-hop reasoning in VQA tasks by integrating decay-aware attention across modalities \cite{zhu2025retention}.
In UAV geolocation, RMT's spatially constrained retention enables robust matching between aerial and satellite views \cite{lin2024single}.  
For image fusion, RetNet facilitates cross-modal shared feature extraction, enabling more coherent integration of text and visual signals \cite{wang2025multi}.

\paragraph{Robotic Perception.}  RAMPGrasp deploys multiscale retention to improve robustness against occlusions and cluttered scenes \cite{huang2025rampgrasp}.
VVNet fuses RetNet modules with ViT backbones to address noise and visibility challenges in underwater imagery \cite{liu2024vvnet}.
HFA-Net employs RMT's MaSA to embed fine-grained spatial priors for subtle facial movement detection \cite{zhang2025hfa}.
The RetNet-based RMT block is integrated into the YOLOv9s backbone network, enhancing local and global feature extraction capabilities \cite{xu2024rmt}.

\subsection{Natural Science}
\label{sec:n_s}
RetNet has demonstrated remarkable versatility across diverse domains in the natural sciences, attributed primarily to its efficient retention mechanism, scalable attention modeling, and capacity to encode long-range dependencies. 

\paragraph{Chemistry.}  \citet{miao2024gr} augmented molecular feature learning by embedding the retention structure during information propagation.
\citet{knitter2024retentive,knitter2024exploration} applied Retenet to Neural-Network Quantum States (NQS).
RetNet's utility extends to lithium-ion battery state-of-health (SoH) estimation, where its retention mechanism excels in capturing temporal degradation patterns \cite{chen2024state}. 

\paragraph{Physics.} JetRetNet exploits retention mechanism to encode multiscale dependencies among tracking and vertex features~\cite{guvenli2024b}.
Radio-frequency signal classification integrates bidirectional retention and cross-block state fusion to accommodate the causal structure of modulation tasks~\cite{han2025radio}.
RAD addresses anomaly detection in cyber-physical systems by employing multi-scale retention and rotational positional encodings to model long-term dependencies efficiently \cite{min2025retentive}.
\citet{cheng2025transrad} exploited RMT enables fine-grained target modeling in radar perception by distributing attention according to spatial proximity.
RetNet has been adapted to assess transient stability through a time-adaptive framework \cite{zhang2024transient}.
In planetary environments, RetNet's retention mechanism has been employed for unmanned aerial vehicle (UAV) monocular visual odometry \cite{liu2025mar}. 

\paragraph{Biology and Medicine.} In biomedical sequence analysis, RetNet have been explored for haplotype assembly \cite{luo2025deephapnet}. \citet{liu2024variation} incorporated RetNet to capture spike protein features. Two RETNets are used to extract drug features and protein features, respectively \cite{peng2024mgndti}. In transcriptomic analysis, RetNet variant processes large-scale cell data \cite{zeng2024cellfm}.

For EEG decoding, RetNet captures bidirectional temporal dependencies \cite{wang2025convolutional}. RetNet denoises EEG signals~\cite{wang2024eegdir}. RetNet decodes temporal patterns for emotion recognition, fused with spatial features \cite{xu2025mitigation}.

In medical image processing,  \citet{elkarazle2023retseg} enabled real-time polyp segmentation with bidirectional retention. \citet{chu2024pfprnet} enhanced polyp segmentation with spatial distance-based retention. RetNet models spatial correlations for echocardiography segmentation \cite{lin2025temporal}. \citet{zhou2024gradient} improved CT denoising via co-retention mechanism. RetNet captures rotation-invariant features for medical image classification \cite{li2025resgdanet}.

\subsection{Social Engineering}
\label{sec:s_e}
RetNet has been extensively adopted in various societal engineering tasks due to its powerful capability to capture long-range dependencies and retain critical information throughout sequential modeling. 

In building change detection, RetNet extracts and preserves spatial features from remote sensing images~\cite{lin2024change}.  In fire detection, RetNet's Local Attention (LA) enhances global feature extraction in YOLO-based models \cite{kim2024domain}. For earthquake early warning, RetNet encodes nonlinear couplings in seismic wave embeddings~\cite{zhang2024fast}. In photovoltaic forecasting, RetNet extracts high-order features from hazy weather data \cite{yang2024short}. In bridge damage assessment, RetNet captures critical features under varying conditions, enhancing anomaly detection \cite{wang2025deep}. For track circuit entity recognition, multi-scale retention (MSR) optimizes long-distance dependency modeling \cite{chen2025named}. In human-robot collaboration, 3DMaSA extends RetNet to predict force and velocity from video sequences \cite{dominguez2024force}. In coal gangue identification, RetNet's retention mechanism optimizes model size and inference speed \cite{zhang2025intelligent}. The application in tea disease detection, using RetNet for spatial modeling \cite{lin2025disease}.

For urban traffic flow prediction, The Temporal Self-Retention (TSR) block to decode time-dependent features extracted by the Temporal Self-Attention module \cite{li5170149adaptive}. Spatial RetNet and temporal RetNet both utilizing multi-scale retention mechanism to effectively capture spatial and temporal dependencies \cite{zhu2024spatial}. Another study by employing RetNet's causal decay mechanism in the temporal branch~\cite{long2024st}. Meanwhile, H-RetNet supports heterogeneous inputs through parallel branches and modality-specific retention strategies~\cite{yan2025multimodal}.
For more applications of RetNet, the reader is referred to \ref{audio}, \ref{others}.




\section{Challenges and Future Directions}
\label{sec:challege_future}
\subsection{Expanding RetNet to Future Applications}
\label{sec:expanding}
In multimodal sentiment analysis, RetNet integrates textual, visual, and auditory signals for accurate emotion recognition. In autonomous systems, it processes real-time LIDAR, camera, and audio streams to enhance decision-making. Future work should focus on improving cross-modal alignment and robustness to noise and resolution variability.

In healthcare, RetNet’s capacity to model high-dimensional longitudinal data enables predictive modeling for personalized medicine. By leveraging data from electronic health records, wearables, and medical imaging, RetNet can forecast disease trajectories such as chronic or neurodegenerative conditions.

\subsection{Technology and Hardware Optimization}
\label{sec:technology}
As RetNet’s applications scale to large-scale deployments, energy efficiency becomes a critical consideration, particularly for edge and mobile environments. The retention mechanism’s reduced computational complexity offers inherent energy savings compared to Transformers, but further optimizations are necessary to meet the demands of sustainable computing.  

RetNet's distinct computational patterns, including parallel, recurrent, and chunkwise recurrent blending, require customized hardware solutions to fully exploit its efficiency. Developing RetNet-specific accelerators, such as retention-aware compute units, can significantly enhance performance.

\subsection{Security and Stability}
\label{sec:security}
Similar to Transformer-based models, RetNet exhibits vulnerabilities to adversarial attacks. Its retention mechanism, while effective for capturing long-term dependencies, may amplify sensitivity to adversarial perturbations, potentially compromising reliability in safety-critical applications such as secure communications or surveillance systems. 

RetNet models, trained on large-scale datasets, often inherit societal biases, such as those related to gender, race, or socioeconomic status, embedded in the training data. These biases can subtly skew predictions, leading to unintended and inequitable outcomes. 
\subsection{Ecosystem and Community Development}
\label{sec:ecd}
The widespread adoption of RetNet, in its capacity as an emerging neural architecture, is contingent on the development of a robust open source ecosystem. The success of architectures such as Transformer, BERT, and GPT, whose proliferation has been significantly supported by accessible and well-maintained open source libraries, has provided a foundation for RetNet. The requirement for an easy-to-use, feature-complete, and extensible software framework to accelerate research and deployment is equally important.

In addition, the establishment of standardized, challenging benchmark datasets and evaluation protocols is critical for objectively assessing RetNet's performance across a range of tasks and domains. Such benchmarks will not only facilitate fair comparisons with existing models, but also ensure the reliability, robustness, and generalizability of RetNet in real-world applications.

\subsection{Challenges in RetNet Enhancement}
\label{sec:cre}
While the explicit decay mechanism in RetNet enables efficient modeling of long-range dependencies by attenuating past information over time, it also introduces limitations in adaptively preserving salient contextual signals. The current decay formulation, typically based on fixed or predefined functions, may not fully capture the dynamic nature of temporal importance across different tasks or modalities. Future research should explore adaptive or learnable decay strategies that allow the model to modulate memory retention based on input characteristics or task demands. For instance, incorporating gating mechanisms or attention-informed decay functions could enable RetNet to selectively preserve or forget past information more effectively. 


\section{Conclusion}
\label{sec:conclusion}
This paper presents a thorough survey of the current literature on RetNet, offering analytical insights, exploring practical applications, and outlining key challenges and future research directions. As the first dedicated review of RetNet to our knowledge, this work seeks to capture the evolving landscape of RetNet-related studies and provide valuable perspectives to support continued innovation and exploration in this emerging field.

\section*{Limitations}
This paper provides a comprehensive review of Retentive Network (RetNet) and its theoretical foundations, empirical performance, and practical applications.
Nevertheless, due to the rapidly evolving nature of this research area, especially in terms of variant designs and scalability optimization, certain notable works may have been inadvertently omitted.
Additionally, a number of studies discussed in this survey rely on earlier benchmarks or smaller-scale evaluations, which may not fully reflect the performance characteristics of RetNet models in large-scale or real-world scenarios. We encourage future research to incorporate state-of-the-art implementations and diverse application contexts to offer more comprehensive and practical guidance.


\bibliography{custom}

\appendix
\section{Audio}
\label{audio}
In recent years, the RetNet has demonstrated significant promise across various audio-related tasks, particularly due to its capacity to efficiently model long-range dependencies while maintaining linear computational complexity. 

\citet{huang2024long} advanced long-form speaker diarization by developing the RetNet-EEND framework, which replaces the Transformer encoder in the EEND-EDA model with a RetNet-based architecture. 
\citet{liang2024ls} proposed the LS-EEND model, which replaces the conventional masked self-attention mechanism in the encoder with Retention, enabling the model to achieve linear time complexity and improved efficiency. 
RetNet has also demonstrated strong potential in the domain of speech enhancement. In the LRetUNet architecture, \citet{zhang2025lretunet} integrated RetNet with LSTM units to construct a time-frequency representation module specifically designed for single-channel speech enhancement.
\section{Others}
\label{others}
RetNet has found versatile applications in multiple domains, leveraging its capability to efficiently model long-term dependencies and handle complex sequential data.

In multimodal representation learning, VL-MFER introduces a bidirectional RetNet (Bi-RetNet) that exploits both parallel and recursive forms of multiscale retention to fuse visual and language modalities \cite{guo2024vl}. 

In the educational domain, a customized Retentive Module extends the original RetNet with a multiscale retention layer, capturing not only the temporal dependencies between student interactions but also their forgetting patterns, thereby enhancing the precision of learning state modeling \cite{linhao2024discovering}.

RetNet has also proven effective for time-series forecasting. LeRet leverages a causal retention encoder alongside multiscale retention modules to enhance nonlinear feature extraction in repaired sequences \cite{huang2024leret}. In reinforcement learning-based recommender systems, RetNet substitutes traditional masked attention with a segmented multiscale retention scheme, significantly improving efficiency and robustness in long-range modeling \cite{wang2024retentive}. For sequential recommendation, dual RetNet modules encode both item and user history, enriching the personalized representation space \cite{wu2024diffusion}. CFPSG further incorporates RetNet into a unified framework to support next-POI prediction by capturing fine-grained temporal correlations \cite{feiyu2024cfpsg}.

RetNet's architectural flexibility makes it well-suited for modeling motion and control dynamics. In EQNet, RetNet serves as the core of a structured state-space model, improving encoding efficiency in multi-agent motion forecasting \cite{huang2025boost}. Similarly, the NC-RetNet introduces a non-causal retention mask to access both past and future frames within blocks, improving 3D human pose estimation while maintaining low latency \cite{zheng20243d}. In TT-DF, RetNet powers the motion-guided branch, balancing long-range dependency modeling and computational cost \cite{yang2024tt}. For robotic manipulation, it is employed to process time-series inputs from learned impedance control dynamics \cite{okada2024contact}.

Language and reasoning tasks also benefit from RetNet's capabilities. In ASTE, a novel bidirectional retention scheme inspired by RetNet bridges sequential and syntactic modeling gaps, boosting sentiment triplet extraction performance \cite{yang2024span}. For world modeling, RetNet supports observation, reward, and termination prediction within REM, and is extended via the POP mechanism to generate observation sequences in parallel during imagination \cite{cohen2024improving}.

System-level advancements further highlight RetNet's efficiency. A high-throughput FPGA inference accelerator incorporates RetNet to maximize hardware utility via dual-mode structure and linear computation \cite{nian202477}. In FlashVideo, RetNet functions as the decoder, with parallel retention for training and autoregressive decoding for inference \cite{lei2023flashvideo}. Sable introduces an encoder-decoder RetNet for MARL with cross-retention and dynamic state resetting to better capture long-term dependencies in online settings \cite{mahjoub2025sableperformantefficientscalable}. In software engineering, RetNet is adopted as the encoder for cloud software code generation from multimodal knowledge \cite{zhang2025cloud}.

In the domain of network analysis, RetNet proves invaluable in both encrypted and anomalous traffic scenarios. RN-ETE extends RetNet by incorporating a multi-resolution self-attention mechanism, enabling bidirectional retention within encrypted traffic encoding for more effective network traffic encryption and analysis \cite{qiu2024rn}. On the other hand, UARC applies RetNet's retention-based reconstruction module to model long-term temporal patterns in network traffic, addressing the challenges of anomaly detection in traffic streams \cite{xie2024uarc}.

In dynamic graph deep learning, \citet{chang2024graph} proposed the Graph Retention Network (GRN) as a unified architecture for deep learning on dynamic graphs.
\end{document}